\pdfoutput=1

\documentclass[11pt]{article}

\usepackage[final]{acl}

\usepackage{times}
\usepackage{latexsym}

\usepackage[T1]{fontenc}

\usepackage[utf8]{inputenc}

\usepackage{microtype}

\usepackage{inconsolata}

\usepackage{graphicx}
\usepackage{amsmath}
\usepackage{amssymb} 
\usepackage{balance}
\usepackage{enumitem} 

\title{Exploring the Potential of Large Language Models for Heterophilic Graphs}

\author{
    Yuxia Wu\textsuperscript{1}$^{*}$,
    Shujie Li\textsuperscript{2}$^{*}$,
    Yuan Fang\textsuperscript{1},
    Chuan Shi\textsuperscript{2}
    \\
    \textsuperscript{1}Singapore Management University,
    \textsuperscript{2}Beijing University of Post and Telecommunication\\
    \text{yieshah2017@gmail.com, shujieli@bupt.edu.cn},
    \text{yfang@smu.edu.sg, shichuan@bupt.edu.cn}
}

\begin{document}
\maketitle

\begingroup
\renewcommand\thefootnote{\relax}
\footnotetext{$^{*}$ Co-first authors with equal contribution.}
\endgroup

\begin{abstract}
Large language models (LLMs) have presented significant opportunities to enhance various machine learning applications, including graph neural networks (GNNs). By leveraging the vast open-world knowledge within LLMs, 
we can more effectively interpret and utilize textual data to better characterize heterophilic graphs, where neighboring nodes often have different labels. However, existing approaches for heterophilic graphs overlook the rich textual data associated with nodes, which could unlock deeper insights into their heterophilic contexts. In this work, we explore the potential of LLMs for modeling heterophilic graphs and propose a novel two-stage framework: LLM-enhanced edge discriminator and LLM-guided edge reweighting. In the first stage, we fine-tune the LLM to better identify homophilic and heterophilic edges based on the textual content of their nodes. In the second stage, we adaptively manage message propagation in GNNs for different edge types based on node features, structures, and heterophilic or homophilic characteristics. To cope with the computational demands when deploying LLMs in practical scenarios, we further explore model distillation techniques to fine-tune smaller, more efficient models that maintain competitive performance. Extensive experiments validate the effectiveness of our framework, demonstrating the feasibility of using LLMs to enhance node classification on heterophilic graphs.
\end{abstract}

\section{Introduction}

Large language models (LLMs) have demonstrated remarkable capabilities across a wide range of applications, from natural language processing \cite{LLM20} to computer vision \cite{wang2024visionllm}, leveraging the extensive open-world knowledge LLMs encode. Inspired by these successes, recent efforts have extended the application of LLMs to the graph domain, particularly in text-attributed graphs where node attributes are composed of textual information \cite{li2023survey, zhang2024graphtranslator}. 
In the context of heterophilic graphs, where connected nodes often exhibit contrasting features or class labels \cite{bo2021beyond, sun2022beyond, liang2024predicting}, LLMs offer a unique opportunity to enhance the understanding of complex semantic relationships between these connected nodes, which remains largely unexplored. 

Existing approaches to addressing heterophily in GNNs typically involve extracting shallow embeddings from textual information, using them as initial node features without fully exploiting their rich semantic content. They can be broadly categorized into two main strategies: non-local neighbor extension and architectural refinement \cite{zheng2022graph, gong2024towards}. The former extends the node's receptive field to include distant, high-order neighbors \cite{abu2019mixhop, songordered} or potential connections \cite{jin2021node, wang2022powerful, zou2023se}, thereby enhancing node representations through a broader scope of information integration. The latter modifies the core functions of GNNs, such as the message aggregation and updating functions, to better suit heterophilic contexts \cite{du2022gbk}. 

In summary, current methods for heterophilic graphs largely overlook the rich textual content associated with the nodes in real-world graphs, which can provide deeper insights into heterophilic contexts. For instance, textual content on hyperlinked webpages can enrich the understanding and prediction of heterophilic links. Traditionally, GNNs employ bag-of-words or shallow embeddings to incorporate textual attributes, which are inadequate for capturing complex semantics. While LLMs \cite{zhao2023survey} have been used to empower GNNs for text-attributed graphs \cite{liu2023towards,li2023survey,yu2024few,maoposition}, existing efforts focus on homophilic graphs, leaving heterophilic graphs largely unexplored. 

In this work, we delve into the potential of LLMs for heterophilic graphs. To the best of our knowledge, this is the first investigation into exploiting LLMs for heterophilic graphs. We aim to bridge the gap between the general capabilities of LLMs and the unique characteristics of heterophilic graphs. Specifically, we aim to address the following research questions.

First, \textit{can LLMs be effectively adapted to characterize and identify heterophilic contexts?} 
As LLMs encompass general open-world knowledge, they can be utilized for the semantic understanding of nodes' textual content.  
However, it is not sufficient to merely extract features from the textual content. 
Unlike homophilic graphs, a key distinction of heterophilic graphs is that edges frequently form between dissimilar nodes. Therefore, distinguishing heterophilic edges from homophilic ones is crucial for subsequent aggregation on graphs.
To address this, we leverage LLMs (or any pre-trained language model in general) to identify heterophilic edges. Specifically, we propose \textbf{LLM-enhanced edge discrimination}, where we fine-tune an LLM using Low-Rank Adaptation (LoRA) \cite{hulora} to discriminate heterophilic and homophilic edges based on a limited amount of ground truth labels. This module focuses on adapting the general semantic capabilities of LLMs to the specific task of predicting heterophily between nodes. The fine-tuned LLM is subsequently used to infer heterophilic edges on the graph to facilitate the integration of heterophilic contexts in the next stage.

Second, \textit{can LLMs effectively guide the fine-grained integration of heterophilic contexts into graph models?}
With respect to a target node, nodes with a potential heterophilic (or homophilic) edge provide valuable heterophilic (or homophilic) contexts for the target node.
Given the diverging characteristics of homophilic and heterophilic contexts, it is important to differentiate them when integrating into a GNN. Specifically, heterophilic contexts of a target node can be identified by our LLM-enhanced edge discrimination. Building on this, we further propose \textbf{LLM-guided edge reweighting} to further aggregate heterophilic and homophilic contexts through GNNs. In this module, we aim to learn adaptive weights for both heterophilic and homophilic edges. These weights are adapted to each edge based on its features, structure, and heterophilic or homophilic characteristics, thereby guiding the fine-grained, edge-sensitive aggregation in GNNs.

Additionally, LLMs often incur high computational costs even for inference, limiting their practical deployment for real-world applications. 
To this end, we further explore model distillation techniques \cite{xu2024survey} to condense the knowledge from fine-tuned LLMs into small language models (SLMs) \cite{schick2021s,li2023symbolic,gu2024minillm,pan2024distilling} to speed up the inference stage required for edge discrimination and reweighting. Specifically, we utilize the LLM well-tuned for edge discrimination to generate high-quality pseudo-labels for heterophilic and homophilic edges. These pseudo-labels supplement the limited ground truth labels, forming an expanded label set that further enables the fine-tuning of SLMs. The fine-tuned SLM then replaces the LLM for conducting inference for the edge discrimination and reweighting, maintaining effectiveness while significantly reducing inference time.

To summarize, we propose a two-stage framework that leverages LLMs for Heterophilic Graph modeling (LLM4HeG). Our main contributions are as follows.
 (1) To the best of our knowledge, this is the first study to explore LLMs specifically for modeling heterophilic graphs. This exploration not only opens new research avenues but also provides valuable insights into the capabilities of LLMs in addressing unique graph characteristics such as heterophily.
    (2) We introduce LLM4HeG, a novel two-stage framework that fine-tunes LLMs 
    to enhance GNNs for heterophilic graphs. The two stages, LLM-enhanced edge discrimination and LLM-guided edge reweighting, accurately identify heterophilic edges and adaptively integrate heterophilic contexts into GNNs, respectively.
    (3) We further investigate the distillation of LLMs fine-tuned for heterophilic edge discrimination into SLMs,
    achieving faster inference time with minimal performance degradation.
    (4) Finally, we conduct extensive experiments on five real-world datasets and demonstrate the effectiveness and efficiency of our work.
 
\begin{figure*}[ht]
	\centering
	\includegraphics[scale=0.7]{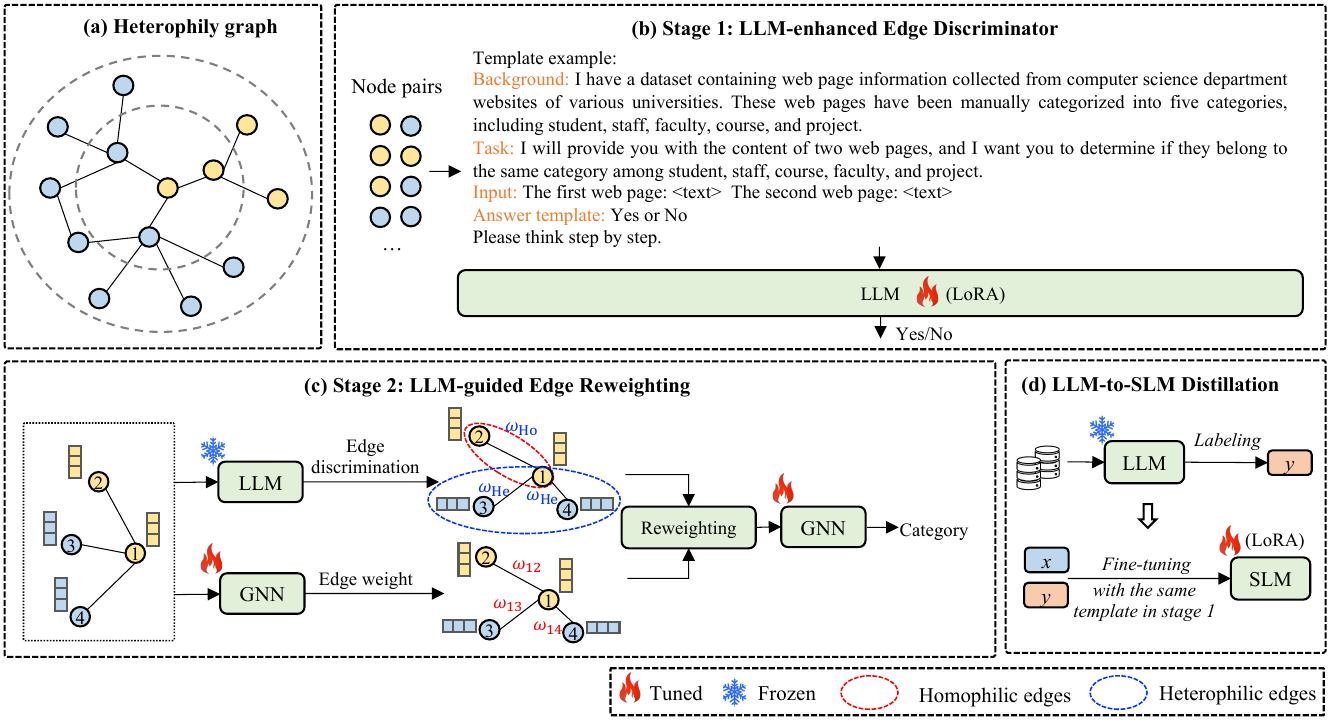}
	\vspace{-0.2cm}
	\caption{Overall framework of the proposed method LLM4HeG.
 }
\label{framework}
\end{figure*}

\section{Related Work}

We review the literature on LLM-based and heterophilic graph learning and highlight the key distinctions of our work from existing studies.

Existing research on LLMs for graph learning includes LLM-based methods that adopt LLM as the backbone and GNN+LLM-based methods that integrate the advantages of both GNNs and LLMs \cite{liu2023towards,chen2024exploring}. The former works focus on aligning graph data with natural language via graph-to-token and graph-to-text approaches \cite{liu2023towards}. The graph-to-token approach involves tokenizing graph data to align it with natural
language, enabling joint understanding with data from other
modalities \cite{zhao2023gimlet,ye2024language}. Graph-to-text focuses on describing graph information using natural language \cite{liu2023evaluating, wang2024can,guo2023gpt4graph}. The latter harnesses the strengths of both language understanding from LLMs
and structural analysis from GNNs by using GNN-centric methods utilizing LLMs to extract node features from
raw data and make predictions using GNNs \cite{heharnessing, xie2023graph} or LLM-centric methods utilizing GNNs to enhance the performance of LLM \cite{tang2024graphgpt,zhang2024graphtranslator}. 

Existing heterophilic graph learning approaches generally fall into two main strategies: non-local neighbor extension and architectural refinement \cite{zheng2022graph,gong2024towards}. Non-local neighbor extension approaches aim to extend the neighbors to include non-local nodes in the graph that may share similar labels or features. These methods often involve high-order neighbor mixing \cite{abu2019mixhop, songordered,yu2024non} or discovering potential neighbors based on various distance measurements, such as feature-based distance \cite{jin2021node, bodnar2022neural}, structure-based distance \cite{peigeom}, or hybrid approaches \cite{wang2022powerful,wang2022powerful2, li2022finding, zou2023se, bi2024make}. Architectural refinement approaches enhance the GNN architecture by employing identifiable message aggregation to discriminate and amplify messages from similar neighbors while minimizing the influence of dissimilar ones \cite{bo2021beyond, zhu2021graph, du2022gbk, liang2024predicting}, or by leveraging inter-layer combinations to capture information from different neighbor ranges \cite{xu2018representation,chienadaptive,zhu2020beyond}, thereby improving the model's representation power in heterophilic graphs. 

The key distinctions of our method lie in two aspects. First, we explore LLMs to enhance text-attributed heterophilic graphs' modeling specifically. While prior works leverage LLMs for text-attributed graphs, including edge reweighting \cite{sun2023large, ling2024link}, they do not explicitly target heterophilic graphs. Additionally, existing heterophilic graph methods often overlook rich textual node attributes, which provide essential semantic contexts. Second, our two-stage framework employs an LLM-enhanced edge discriminator to predict edge types, followed by adaptive message propagation in GNNs using a comprehensive suite of information, including node semantics, structural contexts, and LLM-inferred edge characteristics. While GBK-GNN \cite{du2022gbk} follows a similar two-stage approach, it does not leverage the power of LLMs.

\section{Proposed Model: LLM4HeG}
In this section, we first introduce some preliminaries on the problem formulation and classic GNNs. Then we introduce the overview of the proposed model followed by details of different components.

\subsection{Preliminaries}

\vspace{+0.05cm}\noindent\textbf{Problem formulation.} Let $G=(V,E,X,C)$ denotes a text-attributed graph with a set of nodes $V$ and a set of edges $E$, where each node $v \in V$ is associated with a text document $x_v \in X$. $C$ is the set of node classes. In this paper, we address the task of semi-supervised transductive node classification for heterophilic graphs. Specifically, a subset of the nodes is designated as the training/validation nodes with known class labels, while the goal is to predict the unknown labels of the remaining nodes in the graph.

\vspace{+0.05cm}\noindent\textbf{Classic GNNs.} GNNs typically employ a multi-layer approach to neighborhood aggregation, wherein each node incrementally gathers and aggregates contexts from its neighboring nodes. At the $l^\text{th}$ layer, the representation $\mathbf{h}_v^l \in \mathbb{R}^{d_l}$ of a node $v$ is derived as follows:
\begin{equation}
 \hspace{-1mm}   \mathbf{h}_v^{(l)} = \text{AGGR} ( \mathbf{h}_v^{(l-1)}, \{ \mathbf{h}_u^{(l-1)} : u \in \mathcal{N}(v) \} ),\hspace{-1mm}
\end{equation}
where $d_l$ is the dimension of the node representations at the $l^\text{th}$ layer. The function $\text{AGGR}(\cdot)$ denotes an aggregation mechanism combining the feature vectors of the neighboring nodes, where  $\mathcal{N}(v)$ denotes the set of neighboring nodes of $v$. 

\subsection{Overall Framework}
Fig.~\ref{framework} illustrates the overall framework of our approach LLM4HeG, with a two-stage framework leveraging LLMs for heterophilic graph modeling. 
As shown in Fig.~\ref{framework}(b), Stage 1 involves our LLM-enhanced edge discriminator, where we fine-tune an LLM to distinguish between homophilic and heterophilic edges, utilizing the rich textual data associated with nodes and a limited amount of ground truth label. Following this, in Fig.~\ref{framework}(c), Stage 2 involves LLM-guided edge reweighting to learn adaptive weights for both homophilic and heterophilic edges. These weights are adapted to individual edges by integrating node features, graph structures, and edge types, enabling fine-grained aggregation within GNNs. 

Additionally, to cope with the computational demands of deploying LLMs, we explore a distillation method that condenses the heterophily-specific knowledge of a fine-tuned LLM into a more compact SLM. As shown in Fig.~\ref{framework}(d), we leverage the LLM as a teacher model to generate pseudo labels for additional examples, which can be used to fine-tune an SLM that can perform inference more efficiently without compromising performance.


\subsection{LLM-enhanced Edge Discriminator}
\label{sec:stage1}

In heterophilic graphs, accurately discriminating heterophilic edges from homophilic ones is pivotal for effectively tailoring context aggregation strategies across neighboring nodes. We propose an LLM-enhanced edge discriminator, tapping on the semantic capabilities and open-world knowledge of LLMs beyond conventional shallow feature-based approaches. We first construct the ground truth labels from the training set and then prepare a language template that describes the edge discrimination task for fine-tuning a given LLM. We elaborate on these steps below.

First, to adapt the LLM into an edge discriminator model for heterophilic graphs, we construct ground truth labels to indicate whether a potential edge between two nodes is homophilic or heterophilic. Specifically, if the two nodes have different attributes or categories, their potential relationship is considered heterophilic \cite{yan2022two}. Hence, we select node pairs from the training set and label them as homophilic or heterophilic by comparing their known class labels.
The selection of node pairs depends on the size of the graph. For small graphs, we choose all node pairs; for larger graphs, we choose node pairs within one or two-hop neighborhoods of each other. We will elaborate on the details of node pair selection in Appendix B. Note that we use ``heterophilic/homophilic edge'' to describe a potential relationship between two nodes, even if no explicit edge exists between them.

Next, given such a node pair with a ground-truth label on their homophilic or heterophilic nature, we design a language template to describe the task of heterophilic edge discrimination. We utilize textual information of the node pairs to construct the template as the input text to the LLM, including the \textit{background}, \textit{task}, \textit{input} and \textit{answer template} (Fig.~\ref{framework}(b)). Notably, the \textit{background} also includes a set of node category names specific to the heterophilic graph, which can be regarded as semantic anchors to enhance the LLM's ability to understand the contexts of the given heterophilic graph. 

\vspace{+0.05cm}\noindent\textbf{Fine-tuning.}
The template, together with the ground-truth labels, enables the fine-tuning of LLMs. 
For efficiency, we adopt a parameter-efficient fine-tuning technique called LoRA, which strategically updates only a small fraction of the LLM’s parameters \cite{hulora}. For fine-tuning, we use a typical cross-entropy loss to align the model output with the ground-truth responses:
\begin{equation}
    \mathcal{L}_{\text{fine-tune}} = \textstyle -\frac{1}{N} \sum_{i=1}^{N} \log P_{\theta} (x_i | x_{<i}), \label{ftloss}
\end{equation}
where $N$ is the number of tokens in the sentences, $x_i$ is the $i$-th token to be predicted, and $x_{<i}$ indicates previously generated tokens. $P_{\theta} (x_i | x_{<i}) $ represents the probability of the token $x_i$ given the previous tokens generated by the model.

\vspace{+0.05cm}\noindent\textbf{Inference.} After fine-tuning the LLM for edge discrimination, we can employ it to infer the relationship between any two nodes on the graph, determining whether they have a potentially heterophilic or homophilic edge. During inference, we use the same template to generate input text for each node pair, which is then fed into the fine-tuned LLM to produce a ``\textit{Yes}'' or ``\textit{No}'' answer regarding the edge type.

\subsection{LLM-guided Edge Reweighting}
\label{sec:stage2}
Building on the edge types identified by Stage 1, we proceed to LLM-guided edge reweighting to integrate heterophilic contexts into GNNs. This process leverages the semantic insights acquired from the LLM-enhanced edge discriminator, allowing us to adjust the weight of every individual edge, taking into account various edge-specific factors, including node features and structures, as well as its homophilic or heterophilic nature. 

Specifically, for a node 
$v\in V$, we first extract the representation from the LLM based on the associated textual information:
\begin{equation}
    \mathbf{h}_v^{(0)} = \sigma (\mathtt{LLM}(x_v)\mathbf{W}_e),
\end{equation}
where $x_v$ denotes the raw text of node $v$, $\mathtt{LLM}$ is an LLM encoder, $\mathbf{W}_e$ is a learnable weight matrix and $\sigma$ is the activation function.

For a node $u\in \mathcal{N}_i(v)$, the $i$-hop neighborhood of $v$ \cite{zhu2020beyond}, we infer the edge type of $(u,v)$ using the predictions from Stage 1. Based on the LLM prediction, we formulate an initial weight for $(u,v)$, denoted by $w_{uv}$, as follows.
\begin{align}
  \hspace{-1mm} w_{uv}^{\text{LLM}} &= \begin{cases} 
        \tanh(w_{\text{Ho}}) & \text{if } O_{\text{LLM}}(u,v)=\textit{Yes}, \\
        \tanh(w_{\text{He}}) & \text{if } O_{\text{LLM}}(u,v)=\textit{No},
        \end{cases}\hspace{-1mm}\label{eq:weights} 
\end{align}
where $O_{\text{LLM}}(\cdot)$ represents the output from the LLM-enhanced edge discriminator. The output ``\textit{Yes}'' implies a homophilic edge, whereas ``\textit{No}'' indicates a heterophilic edge. 
This initial LLM-based weight $w_{uv}^{\text{LLM}}$ is defined based on two learnable parameters: $w_{\text{Ho}}$ for homophilic edges and $w_{\text{He}}$ for heterophilic edges. 
%
The two parameters are crucial for modulating the strength and influence of each edge type within the graph. For homophilic edges, where stronger connectivity is often beneficial, the learned parameter may increase the weight, thus amplifying the coherence and communication within similar node clusters. Conversely, for heterophilic edges, which often bridge diverse node groups, the parameter might be adjusted to achieve a balance between reducing noises and maintaining critical cross-group information.

On the other hand, graph-based information, including node features and structures, also provides crucial insights on determining the edge weight. 
While our framework LLM4HeG is designed to be flexible, allowing for the adoption of various GNN backbones, we showcase FAGCN \cite{bo2021beyond} as an example of how our framework can be effectively applied. 
This method learns the edge-specific aggregation weight via a self-gating mechanism:
\begin{equation}
 w_{uv}^{\text{G}} = \tanh\left(\mathbf{g}^{\top} \left[ \mathbf{h}_u \parallel \mathbf{h}_v \right]\right), 
\end{equation}
where $\parallel$ denotes the concatenation operation, and $\mathbf{g}$ is a linear layer to map the concatenated feature into a scalar value, which can be seen as a shared convolutional kernel \cite{velivckovic2018graph}.

Finally, we integrate the LLM-based weight $w_{uv}^{\text{LLM}}$ with the graph-based weight $w_{uv}^{\text{G}}$. While there are many ways to achieve this, we use a simple yet effective method that takes the average of the two weights as the final weight, denoted by $w_{uv}$, as follows.
\begin{align}
w_{uv} &= \textstyle \frac{1}{2}\left(w_{uv}^{\text{LLM}} + w_{uv}^{\text{G}}\right). \label{eq:reweight}
\end{align}
The edge-specific weights further enable fine-grained context aggregation, as detailed below. 
\begin{align}
\mathbf{h}_v^{(l)} &= \textstyle \epsilon \mathbf{h}_v^{(0)} + \sum_{u \in \mathcal{N}_i(v)} \frac{w_{uv}}{\sqrt{d_u d_v}} \mathbf{h}_u^{(l-1)}, \\
\mathbf{h}_{\text{out}} &= \mathbf{W}_o \mathbf{h}_v^{(L)},
\end{align}
where $\epsilon$ is a scaling hyper-parameter, $d_{v} = |N_i(v)|$, $\mathbf{W}_o$ is a weight matrix, and $L$ is the total number of GNN layers.

\vspace{+0.05cm}\noindent\textbf{Training.}
We adopt a typical cross-entropy (CE) loss to train node classification. Additionally, to ensure the learned weights $w_{\text{Ho}}$ and $w_{\text{He}}$ show sufficient separation, we introduce a margin-based regularization term into the CE loss.
\begin{equation}
   \mathcal{L} = \mathtt{CE}(\hat{y}, y) + \lambda \max(0, w_{\text{He}}-w_{\text{Ho}}+\alpha), \label{loss}
\end{equation}
where $\mathtt{CE}(\hat{y}, y)$ represents the cross-entropy loss between the predicted node label $\hat{y}$ and the ground-truth label $y$, $\lambda$ controls the influence of the regularization term and $\alpha$ represents the margin that enforces a minimum difference between the weights $w_{\text{Ho}}$ and $w_{\text{He}}$.
During training, we freeze the parameters of the LLM and only update the weights of the GNN model.

\vspace{+0.05cm}\noindent\textbf{Inference.}
During the inference phase of Stage 2, we apply the LLM-enhanced edge discriminator fine-tuned in Stage 1 to generate the edge type for each node pair involved in the test set. Subsequently, we can calculate the output representations for the test nodes based on the reweighted edges for node classification. 

\subsection{LLMs-to-SLMs Distillation}
\label{sec:distill}
In real-world applications, deploying LLMs as edge discriminators introduces substantial computational challenges, even when only the inference phase is required for predicting the edge types on test graphs. 
To mitigate the computational burden of the inference phase, we explore knowledge distillation techniques to transfer the heterophily-specific capabilities of the fine-tuned LLM into more lightweight SLMs \cite{xu2024survey}. 

As shown in Fig.~~\ref{framework}(d), after fine-tuning the LLM in Stage 1, we use it as a teacher model to generate the pseudo-labels for additional node pairs sampled from the entire graph. The labeling process follows the inference phase introduced in Stage 1, asking the LLM whether a given node pair is homophilic or heterophilic based on the input template. These pseudo-labels are combined with the ground-truth labels, forming an expanded label set which is subsequently used to fine-tune the SLMs. The fine-tuning of the SLMs follows the same approach as fine-tuning the LLM in Stage 1, using the same template and the LoRA technique.
Finally, we replace the fine-tuned LLM with the fine-tuned SLM during inference, which predicts the homophilic or heterophilic relationship between any two given nodes to guide edge reweighting in Stage 2.

\vspace{+0.05cm}\noindent\textbf{Learning objectives.} 
We outline the objectives of knowledge distillation in the two stages. In Stage 1, the fine-tuning of SLMs follows the same approach as fine-tuning the LLM shown in Eq.~\eqref{ftloss}. In Stage 2, we train the edge reweighting using the same cross-entropy objective shown in Eq.~\eqref{loss}.


\section{Experiment}

In this section, we present an empirical study to demonstrate the feasibility of leveraging LLMs for node classification on heterophilic graphs. 

\subsection{Experimental Setup}
\vspace{+0.05cm}\noindent\textbf{Datasets.} 
Given that the datasets commonly employed in heterophily graph tasks lack original textual information, we collect publicly available raw text directly from the original data providers and preprocess these datasets. Consequently, our experiments only include datasets that contain raw text.
Specifically, we have prepared five datasets:
Cornell, Texas, and Wisconsin \cite{peigeom}, Actor \cite{tang2009social} and Amazon \cite{platonovcritical}. The Cornell, Texas, and Wisconsin datasets \cite{peigeom} are derived from computer science department websites where webpages serve as nodes and hyperlinks as edges. The Actor dataset \cite{tang2009social} is an actor co-occurrence network with nodes representing actors and edges denoting their collaborations. The Amazon dataset \cite{platonovcritical} is a product co-purchasing network, where nodes represent products and edges link products frequently bought together. The statistical details of the datasets are presented in Table \ref{dataset} with more details provided in Appendix \ref{app:data}. Additionally, we include the edge homophily score $\mathcal{H}(G)$ \cite{yan2022two} for each dataset, which quantifies the level of homophily (1 means perfectly homophily while 0 stands for total heterophily). 

\begin{table}[t]
\centering
\small
\begin{tabular}{lcccc}
\hline
Dataset   & Classes & Nodes  & Edges  & $\mathcal{H}(G)$ \\ 
\hline
Cornell   & 5       & 195    & 304   & 0.13 \\
Texas     & 5       & 187    & 328   & 0.12 \\
Wisconsin & 5       & 265    & 530   & 0.20 \\
Actor     & 5       & 4,416  & 12,172  & 0.56 \\
Amazon    & 5       & 24,492 & 93,050  & 0.38 \\
\hline
\end{tabular}
\vspace{-0.2cm}
\caption{Dataset statistics.}
\label{dataset}
\vspace{-0.3cm}
\end{table}

\vspace{+0.05cm}\noindent\textbf{Baselines.} 
We compare our method against a set of baseline models that fall into two main categories: classic GNN models and heterophily-specific models. The classic GNN models include GCN \cite{kipf2016semi}, GraphSAGE \cite{hamilton2017inductive} and GAT \cite{velivckovic2018graph}. The heterophily-specific models include H2GCN \cite{zhu2020beyond}, FAGCN \cite{bo2021beyond}, JacobiConv  \cite{wang2022powerful}, GBK-GNN \cite{du2022gbk}, OGNN \cite{songordered}, SEGSL \cite{zou2023se}, DisamGCL \cite{zhao2024disambiguated} with more details provided in Appendix \ref{app:baselines}.   

\vspace{+0.05cm}\noindent\textbf{Implementation Details.}
We adopt Vicuna-v1.5-7B \cite{zheng2024judging} as the LLM model and the Bloom model \cite{le2023bloom} with 560M and 1B parameters as SLMs.
Consistent with most existing works \cite{zhu2020beyond}, we randomly split the nodes into train, validation and test sets with a proportion of 48\%/32\%/20\% for Cornell, Texas, Wisconsin and Actor datasets. We set the proportion of Amazon dataset as  50\%/25\%/25\% following \citet{platonovcritical}. In our main experiments, we use FAGCN \cite{bo2021beyond} as the GNN backbone, adhering to the parameter settings outlined in FAGCN. We only fine-tune the other hyperparameters based on the performance observed on the validation sets. All experiments are repeated 10 times, and we report the averaged results with standard deviation. To ensure a fair comparison, we use the initial node features derived from the Vicuna 7B model for all methods. More implementation details and hyper-parameters are introduced in Appendix \ref{app:implem} and \ref{app:hyper}.
\subsection{Performance Comparison}
\begin{table}[t]
\centering
\small
\addtolength{\tabcolsep}{-4pt}
\renewcommand*{\arraystretch}{1.2}
\resizebox{0.48\textwidth}{!}{
    \begin{tabular}{p{1.7cm}cccccc}
    \hline 
     Methods & Cornell & Texas & Wisconsin & Actor & Amazon \\ 
    \hline 
    \multicolumn{6}{c}{\textit{Classic GNNs}} \\
    \hline 
    GCN         & 52.86\scriptsize{±1.8} & 43.64\scriptsize{±3.3} & 41.40\scriptsize{±1.8} & 66.70\scriptsize{±1.3} & 39.33\scriptsize{±1.0} \\
    GraphSAGE   & 75.71\scriptsize{±1.8} & 81.82\scriptsize{±2.5} & 80.35\scriptsize{±1.3} & 70.37\scriptsize{±0.1} & 46.63\scriptsize{±0.1} \\ 
    GAT        & 54.28\scriptsize{±5.1} & 51.36\scriptsize{±2.3} & 50.53\scriptsize{±1.7} & 63.74\scriptsize{±6.7} & 35.12\scriptsize{±6.4} \\
    \hline 
    \multicolumn{6}{c}{\textit{Heterophily-specific GNNs}} \\
    \hline 
    H2GCN       & 69.76\scriptsize{±3.0} & 79.09\scriptsize{±3.5} & 80.18\scriptsize{±1.9} & 70.73\scriptsize{±0.9} & 47.09\scriptsize{±0.3} \\
    FAGCN       & 76.43\scriptsize{±3.1} & 84.55\scriptsize{±4.8} & 83.16\scriptsize{±1.4} & 75.58\scriptsize{±0.5} & 49.83\scriptsize{±0.6} \\
    JacobiConv  & 73.57\scriptsize{±4.3} & 81.80\scriptsize{±4.1} & 76.31\scriptsize{±11.3} & 73.81\scriptsize{±0.3} & 49.43\scriptsize{±0.5} \\
    GBK-GNN & 66.19\scriptsize{±2.8} &80.00\scriptsize{±3.0} &72.98\scriptsize{±3.3} &72.49\scriptsize{±1.0} &44.90\scriptsize{±0.3} \\
    OGNN   & 71.91\scriptsize{±1.8} & 85.00\scriptsize{±2.3} & 79.30\scriptsize{±2.1} & 72.08\scriptsize{±2.4} & 47.79\scriptsize{±1.6} \\
    SEGSL   & 66.67\scriptsize{±4.1} & 85.00\scriptsize{±2.0} & 79.30\scriptsize{±1.8} & 72.73\scriptsize{±0.8} & 47.38\scriptsize{±0.2} \\
    DisamGCL    & 50.48\scriptsize{±2.0} & 65.00\scriptsize{±1.2} & 57.89\scriptsize{±0.0} & 67.78\scriptsize{±0.3} & 43.90\scriptsize{±0.4} \\
    
    \hline 
    \multicolumn{6}{c}{\textit{LLM4HeG} (fine-tuned LLM/SLMs and distilled SLMs )} \\
    \hline 
      Vicuna 7B      & \textbf{77.62}\scriptsize{±2.9} & \textbf{89.09}\scriptsize{±3.3} & 86.14\scriptsize{±2.1} & \textbf{76.82}\scriptsize{±0.5} & 51.53\scriptsize{±0.4} \\
    
      Bloom 560M     & 75.48\scriptsize{±2.1} & 80.00\scriptsize{±4.0} & \underline{86.49}\scriptsize{±1.9} & \underline{76.16}\scriptsize{±0.6} & 51.52\scriptsize{±0.5} \\
    
      Bloom 1B     & 75.71\scriptsize{±1.4} & 83.86\scriptsize{±2.8} & 83.86\scriptsize{±1.7} & 74.99\scriptsize{±0.5} & \textbf{52.33}\scriptsize{±0.6} \\
    
      7B-to-560M    & 75.00\scriptsize{±4.0} & \underline{88.18}\scriptsize{±2.2} & \textbf{87.19}\scriptsize{±2.5} & 75.78\scriptsize{±0.2} & 51.51\scriptsize{±0.4} \\
    		
      7B-to-1B    & \underline{77.38}\scriptsize{±2.7} & \underline{88.18}\scriptsize{±4.0} & 86.14\scriptsize{±1.5} & 75.37\scriptsize{±0.9} & \underline{51.58}\scriptsize{±0.4} \\
    		
    \hline 
    \end{tabular}
}
\vspace{-0.2cm}
\caption{Accuracy for node classification of different methods. (Best results bolded; runners-up underlined.)}
\label{result}
\end{table}

Table \ref{result} shows the average accuracy and the standard deviation of the baselines and LLM4HeG with fine-tuned LLM/SLMs and distilled SLMs. For LLM4HeG, we choose the Vicuna 7B as the LLM and the Bloom models with 560M and 1B as SLMs. We conduct experiments with various strategies for the LLM and SLMs to explore their performance. (1) ``Vicuna 7B'': We fine-tune the LLM via LoRA technique for edge discrimination. (2) ``Bloom 560M'' and ``Bloom 1B'': We directly fine-tune SLMs via LoRA in the same way as ``Vicuna 7B''. (3) ``7B-to-560M'' and ``7B-to-1B'': We distill the Vicuna 7B model to Bloom 560M and 1B models, respectively, as introduced in Sect.~\ref{sec:distill}.

Among the baselines, heterophily-specific GNNs generally outperform classic GNNs, while our methods consistently achieve the best performance. These results indicate that LLM4HeG effectively captures the complex relationships among different nodes in heterophilic graphs. 

For different strategies of LLM4HeG, directly fine-tuning the SLMs often leads to notable performance decline compared to fine-tuning the LLM. However, the distilled SLMs attain performance comparable to that of the fine-tuned LLM, demonstrating the effectiveness of the distillation process in retaining heterophily-specific knowledge. The only exception is the performance on the Amazon dataset, where the directly fine-tuned SLMs achieve similar performance as the LLM and the distilled ones. This may be due to the semi-structured patterns within the textual descriptions in the Amazon dataset such as product specifications, which are relatively easy to capture by both LLMs and SLMs. We also present experiments that directly utilize LLMs for node classification in Appendix \ref{app:LLM}.

\subsection{Model Analysis}

\noindent\textbf{Ablation Study.} To evaluate the effectiveness of the learnable weight for adaptive message passing and deep node features from LLM, we conducted experiments using different variants of our model: (1) \textit{w/o reweight}: This variant only uses the graph-based weight $w_{uv}^G$ (2) \textit{w/o learnable weights}: This variant uses fixed weights instead of learnable ones. Specifically, we manually set the weights for homophilic and heterophilic pairs to 1 and -1, respectively, as these values were found to perform well across most datasets.

\begin{figure}[t]
	\centering
	\includegraphics[scale=0.7]{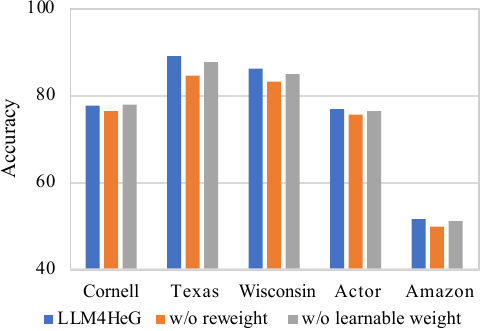}
 \vspace{-0.2cm}
	\caption{The effectiveness of learnable weight. 
 }
\label{ablation}
\end{figure}

 As shown in Fig.~\ref{ablation}, the performance drops after removing the reweighting mechanism. When reweighting is applied, the \textit{w/o learnable weight} variant outperforms the backbone model and the learnable weight of LLM4HeG shows the best performance. These results indicate the significance of learnable adaptive reweighting guided by LLM for message aggregation in GNNs. 

\begin{table}[t]
\centering
\small
\addtolength{\tabcolsep}{-4pt}
\renewcommand*{\arraystretch}{1.1}
\resizebox{0.48\textwidth}{!}{
    \begin{tabular}{@{}ccccccc}
    \hline 
    Model  & Cornell& Texas & Wisconsin & Actor & Amazon & Average
    \\ \hline 
     Vicuna 7B & 65.71 & 64.00 & 92.66 & 81.50 & 44.68 & 69.71  \\ \hline 
     Bloom 560M & 47.62 & 26.51 & 71.62 & 79.02 & 56.26 & 56.21  \\ 
        Bloom 1B & 40.86 & 23.91 & 79.76 & 79.52 & 59.89 & 56.78  \\
     \hline 
     7B-to-560M & 50.85 & 64.86 & 80.75 & 81.03 & 50.77 & 65.65  \\ 
      7B-to-1B & 51.72 & 80.00 & 75.95 & 80.47 & 51.48 & 67.92  \\ 
     \hline 
    \end{tabular}
}
\vspace{-0.2cm}
\caption{F1 scores for edge discrimination of fine-tuned LLM/SLMs and distilled SLMs.}
\label{LLM-stage1}
\vspace{-0.3cm}
\end{table}

\begin{figure}[t]
\centering
\includegraphics[scale=0.7]{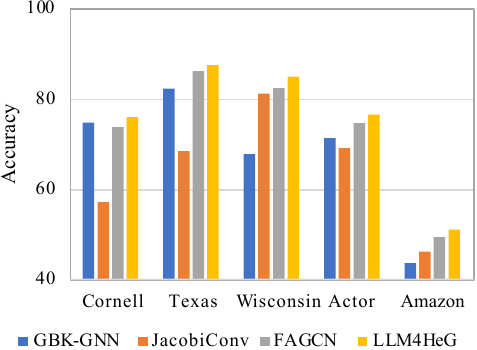}
 \vspace{-0.2cm}
\caption{The accuracy of inductive node classification. 
 }
\label{inductive}
\vspace{-0.2cm}
\end{figure}

\begin{figure*}[t]
	\centering
	\includegraphics[scale=0.6]{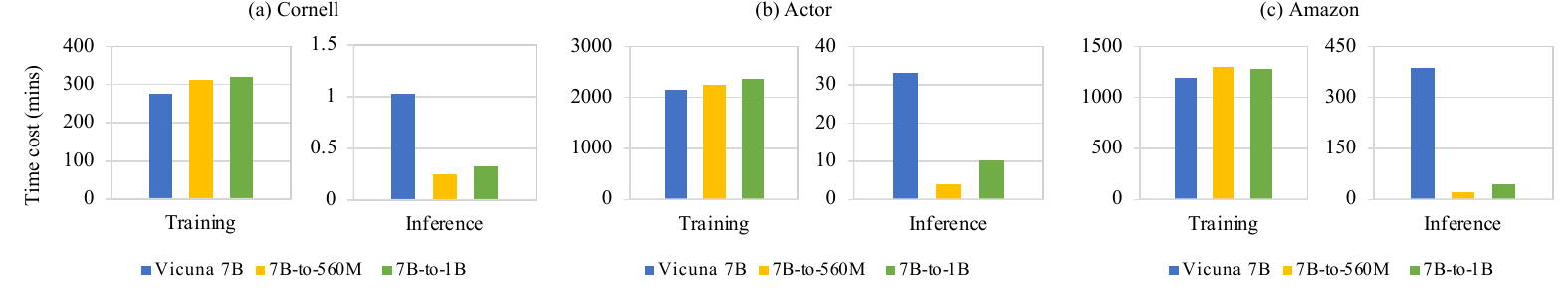}
	\vspace{-0.7cm}
	\caption{Analysis on the efficiency of the fine-tuned LLM and distilled SLMs.
 }
\label{time}
\end{figure*}

\vspace{+0.05cm}\noindent\textbf{Analysis of edge discrimination by LLM/SLMs.} As the edge reweighting in Stage 2 depends on the effectiveness of edge discrimination in Stage 1, 
we further evaluate the edge classification performance (F1 score) of different models on the node pairs used for Stage 2. As shown in Table~\ref{LLM-stage1}, compared with the fine-tuned LLM, directly fine-tuning the SLMs generally gives worse edge discrimination performance, which is reasonable considering the model capacity. However, the distilled SLM manages to maintain a comparable performance with only a marginal drop. This indicates the effectiveness of the distillation process, which allows the distilled SLM to retain heterophily-specific knowledge of the fine-tuned LLM. The results are consistent with the node classification performance reported in Table~\ref{result}.

\vspace{+0.05cm}\noindent\textbf{Performance on inductive setting.} Our method is inherently adaptable to inductive node classification, where test nodes remain entirely unseen during training. We conduct additional experiments on inductive test sets, requiring the model to classify them based on their raw textual attributes and structural connections. As shown in Fig.~\ref{inductive}, our approach consistently outperforms baseline methods in this setting. These results indicate that our method can effectively benefit from the semantic understanding of LLMs in both transductive and inductive settings.

\vspace{+0.05cm}\noindent\textbf{Efficiency study.}
Fig.~\ref{time} illustrates the efficiency of our framework using an LLM or distilled SLMs in terms of training and inference times, measured in minutes. For the LLM approach, the training time denotes the fine-tuning time in Stage 1 on the selected node pairs (see Section~\ref{sec:stage1}). For the distilled SLMs, as discussed in Section~\ref{sec:distill}, we employ the fine-tuned LLM to generate pseudo-labels for further fine-tuning the SLM. Thus, its training time includes fine-tuning the LLM, generating the pseudo-labels, and fine-tuning the SLM. Thus, the total training time for model distillation is slightly higher than the LLM. On the other hand, inference time includes predicting edge types using either the LLM or distilled SLM, as required by the edge reweighting module in Stage 2. It is worth noting that the inference times of SLMs are significantly lower than LLMs, especially for larger datasets such as Amazon. Hence, the distilled SLMs can be more easily deployed given their smaller size and faster inference time, while maintaining competitive performance as shown in Table~\ref{result}.

\subsubsection{Plug-and-play with various backbones}
Our method is designed to be highly flexible, allowing it to be integrated with various GNN backbones. To demonstrate the flexibility, we tested our approach with several well-known GNN architectures, including GCN, GAT, H2GCN, FAGCN and GBK-GNN. We provide the implementation details of the integration process in Appendix \ref{app:implem}. Table \ref{backbone} indicates that our method consistently enhances the performance of these backbones. This capability highlights the versatility of our approach, making it valuable for improving multiple backbones without the need for significant modifications.

\begin{table}[t]
\centering
\small
\addtolength{\tabcolsep}{-3.5pt}
\renewcommand*{\arraystretch}{1.1}
\resizebox{0.48\textwidth}{!}{
\begin{tabular}{@{}lccccc@{}}
\hline 
 & Cornell & Texas & Wisconsin & Actor & Amazon \\ 
\hline 
GCN & 52.86\scriptsize{±1.8} & 43.64\scriptsize{±3.3} & 41.40\scriptsize{±1.8} & 66.70\scriptsize{±1.3} & 39.33\scriptsize{±1.0} \\
$\text{+LLM4HeG}$       & 66.19\scriptsize{±1.0} & 68.18\scriptsize{±2.0} & 76.84\scriptsize{±2.6} & 71.68\scriptsize{±1.0} & 40.98\scriptsize{±0.7} \\
\hline 
GAT     & 54.28\scriptsize{±5.1} & 51.36\scriptsize{±2.3} & 50.53\scriptsize{±1.7} & 63.74\scriptsize{±6.7} & 35.12\scriptsize{±6.4} \\ 
$\text{+LLM4HeG}$      & 58.57\scriptsize{±4.9} & 58.18\scriptsize{±2.3} & 57.54\scriptsize{±6.1} & 70.78\scriptsize{±0.7} & 36.01\scriptsize{±5.8} \\ 
\hline 
H2GCN     & 69.76\scriptsize{±3.0} & 79.09\scriptsize{±3.5} & 80.18\scriptsize{±1.9} & 70.73\scriptsize{±0.9} & 47.09\scriptsize{±0.3} \\ 
$\text{+LLM4HeG}$      & 76.43\scriptsize{±3.6} & 84.77\scriptsize{±1.0} & 86.49\scriptsize{±1.1} & 74.51\scriptsize{±0.6} & 52.14\scriptsize{±0.4} \\ 
\hline 
FAGCN     & 76.43\scriptsize{±3.1} & 84.55\scriptsize{±4.8} & 83.16\scriptsize{±1.4} & 75.58\scriptsize{±0.5} & 49.83\scriptsize{±0.6} \\ 
$\text{+LLM4HeG}$      & 77.62\scriptsize{±2.9} & 89.09\scriptsize{±3.3} & 86.14\scriptsize{±2.1} & 76.82\scriptsize{±0.5} & 51.53\scriptsize{±0.4} \\ 
\hline 
GBK-GNN  & 66.19\scriptsize{±2.8} &80.00\scriptsize{±3.0} &72.98\scriptsize{±3.3} &72.49\scriptsize{±1.0} &44.90\scriptsize{±0.3} \\
$\text{+LLM4HeG}$     &68.57\scriptsize{±2.6}
&81.82\scriptsize{±2.0}
&76.14\scriptsize{±1.4}
&73.39\scriptsize{±0.6}
&48.25\scriptsize{±0.3} \\ 

\hline 
\end{tabular}
}
\vspace{-0.2cm}
\caption{The accuracy for node classification of LLM4HeG with different backbones.}
\label{backbone}
\end{table}

\subsubsection{Hyper-parameter study.}
Finally, we analyze the impact of a key hyper-parameter in our work, namely the weight margin $\alpha$ in Eq.~\eqref{loss}, the training loss of Stage 2. We vary it over $\{0, 0.1, 0.3, 0.5, 0.7, 0.9\}$, and report the results for three datasets in Fig.~\ref{hyper: alpha}, with the results for the remaining datasets in Appendix \ref{app:hyper}. Generally speaking, when $\alpha$ is too low or too high, the performance tends to drop, suggesting that a balance is required to appropriately consider the difference between the learned weights. 
In particular, $[0.3,0.5]$ appears to be a good range for $\alpha$ on these datasets.
\begin{figure}[t] 
\centering
    \centering
    \includegraphics[scale=0.53]{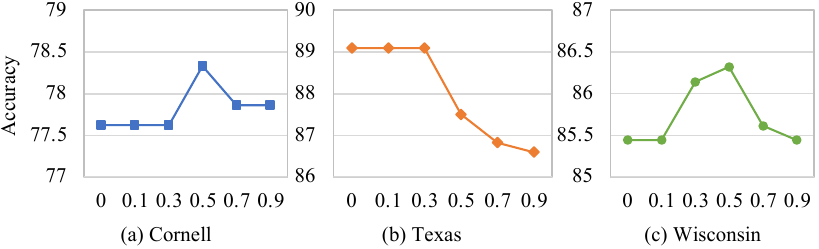}
\vspace{-0.2cm}
\caption{Effect of the edge weight margin $\alpha$.}
\label{hyper: alpha}
\vspace{-0.3cm}
\end{figure}

We also conduct experiments on the influence of the initial values of $w_{\text{Ho}}$ and $w_{\text{He}}$ in Eq.~\eqref{eq:weights}. The results in Table~\ref{tab:initial-weights} show that the classification performance is relatively stable across different initial values of $w_{\text{Ho}}$ and $w_{\text{He}}$, with minor variations observed for specific datasets. This suggests that the model is robust to different initializations, and can generally find the optimal values for $w_{\text{Ho}}$ and $w_{\text{He}}$  despite different initializations.

\begin{table}[h]
    \centering
\addtolength{\tabcolsep}{-4pt}
\renewcommand*{\arraystretch}{1.2}
\resizebox{0.48\textwidth}{!}{    
    \begin{tabular}{cc|ccccc}
        \hline
        $w_{\text{Ho}}$ & $w_{\text{He}}$ & Cornell & Texas & Wisconsin & Actor & Amazon \\
        \hline
        0.5  & 0.5  & 78.57\scriptsize{±2.6} & 89.55\scriptsize{±2.9} & 85.44\scriptsize{±2.5} & 76.83\scriptsize{±0.6} & 51.47\scriptsize{±0.4} \\ \hline
        1    & 0    & 77.86\scriptsize{±2.4} & 89.55\scriptsize{±2.5} & 85.44\scriptsize{±2.1} & 77.00\scriptsize{±0.4} & 51.42\scriptsize{±0.6} \\ \hline
        1.5  & -0.5 & 77.62\scriptsize{±2.9} & 89.09\scriptsize{±3.3} & 86.14\scriptsize{±2.1} & 76.82\scriptsize{±0.5} & 51.53\scriptsize{±0.4} \\ \hline
        2    & -1   & 78.57\scriptsize{±1.8} & 87.73\scriptsize{±3.6} & 85.44\scriptsize{±1.8} & 76.88\scriptsize{±0.5} & 51.40\scriptsize{±0.4} \\ \hline
        2.5  & -1.5 & 78.57\scriptsize{±1.5} & 87.95\scriptsize{±2.3} & 86.14\scriptsize{±1.8} & 76.99\scriptsize{±0.5} & 51.30\scriptsize{±0.3} \\ 
        \hline
    \end{tabular}
    }
    \vspace{-0.2cm}
    \caption{Effect of the initial weights of $w_{\text{Ho}}$ and $w_{\text{He}}$.}
    \label{tab:initial-weights}
\end{table}

\section{Conclusion}
In this study, we explored the potential of LLMs to enhance the performance of GNNs for node classification on heterophilic graphs. We introduced a novel two-stage framework LLM4HeG, integrating LLMs into the GNN learning process through an LLM-enhanced edge discriminator and an LLM-guided edge reweighting module. LLM4HeG allows more precise identification of heterophilic edges and finer-grained context aggregation, leveraging the rich semantics in nodes' textual data. Additionally, to address the computational challenges of deploying LLMs, we implemented model distillation techniques to create smaller models that achieve much faster inference while maintaining competitive performance. Our extensive experiments demonstrate that LLM4HeG significantly improves node classification on heterophilic graphs, underscoring the potential of LLMs for advancing complex graph learning.

\section*{Limitations}
Despite the promising results obtained by our approach LLM4HeG, it is important to acknowledge several limitations. (1) LLM4HeG follows a two-stage pipeline, which may lead to error accumulation between stages, as compared to end-to-end approaches that jointly optimize the entire process. (2) The effectiveness of LLM4HeG  depends on the availability and quality of textual data associated with nodes, particularly the alignment between the semantics of textual attributes and the class labels. The model may struggle to deliver optimal results when textual data is sparse or irrelevant. (3) LLM4HeG operates in a supervised learning paradigm, requiring labeled data for training, which can limit its scalability and applicability in domains where labeled data is scarce or costly to obtain.

\section*{Acknowledgments}
This research / project is supported by the Ministry of Education, Singapore, under its Academic Research Fund Tier 2 (Proposal ID: T2EP20122-0041). Any opinions, findings and conclusions or recommendations expressed in this material are those of the author(s) and
do not reflect the views of the Ministry of Education, Singapore.

\bibliography{ref}

\appendix
\section*{Appendices}

\section{More Details for Datasets}
\label{app:data}
\begin{itemize}[leftmargin=*]
    \item Cornell, Texas, and Wisconsin \cite{peigeom} are 
    collected from computer science departments at various universities. In these datasets, each node corresponds to a web page, while edges represent hyperlinks connecting these pages. In our experiments, we use the original webpage data as the textual information for each node. 
    
    \item Actor \cite{peigeom,tang2009social} is an actor-only induced subgraph of the film-director-actor-writer network. 
    In this graph, nodes represent actors, and an edge between two nodes indicates their co-occurrence on the same Wikipedia page. The task involves categorizing actors into five distinct classes based on their roles. We selected the actors based on category information provided in the metadata, focusing on those with high occurrence frequencies. The category keywords of the selected actors include ``American film actors", ``American film and television actors", ``American stage and television actors",  ``English" and ``Canadian". Afterward, we construct the graph based on the edges and remove the isolated nodes from the graph.
    
    \item Amazon \cite{platonovcritical} is constructed from the Amazon product co-purchasing network metadata. 
    In this dataset, nodes represent products such as books, music CDs, DVDs, and VHS video tapes. Edges link products that are frequently bought together. The goal is to predict the average rating a product receives from reviewers, with ratings grouped into five classes. To manage the graph's complexity, only the largest connected component of the 5-core of the graph is considered.
\end{itemize}

\section{More Implementation Details}
\label{app:implem}

\vspace{+0.05cm}\noindent\textbf{Node Pairs.}
In Stage 1, we train the edge discriminator by sampling node pairs from the graph, with the selection process tailored to the characteristics of each dataset. For small graphs like Cornell, Texas, and Wisconsin, we select all node pairs within the training set, including those without direct edges. For the Actor dataset, node pairs are selected based on the 1-hop and 2-hop neighbor relationship, while for the Amazon dataset, we focus on node pairs with a 1-hop neighbor relationship, taking into account the graph's size.

In the distillation process, we use node pairs from the validation and testing data as additional samples, allowing the fine-tuned LLM to generate pseudo-labels. For small graphs like Cornell, Texas, and Wisconsin, we select node pairs with 1-hop and 2-hop neighbor relationships, while for the Actor and Amazon datasets, we choose node pairs with 1-hop neighbor relationships. The number of node pairs used for training in Stage 1 and distillation is shown in Table \ref{app: node pairs}.

\begin{table}[tbp]
\centering
\small
\setlength{\tabcolsep}{2.5pt}
\begin{tabular}{lccccc}
\hline
Dataset       & Cornell & Texas & Wisconsin & Actor & Amazon \\
\hline
Training      & 4,186    & 3,741  & 7,626      & 36,248 & 23,210 \\
Distillation$\star$  & 916     & 991   & 1,299      & 1,781  & 11,422 \\
\hline
\multicolumn{6}{l}{}\\[-3mm]
\multicolumn{6}{l}{\scriptsize $\star$: the number of additional samples for distillation.}
\end{tabular}
\vspace{-0.2cm}
\caption{The number of node pairs in Stage 1 and the distillation process.}
\label{app: node pairs}
\end{table}

\vspace{+0.05cm}\noindent\textbf{Backbones.}
We provide the implementation details for integrating the LLM4HeG with other backbones. As discussed in Sec. \ref{sec:stage2}, the edge reweighting in Stage 2 combines the LLM-based weight with graph-based information. For the backbones that don't contain an additional edge weight learning module (\textit{e.g.}, GCN \cite{kipf2016semi} and H2GCN \cite{zhu2020beyond}), we only use the edge weight obtained from LLM. For the backbones with specific designs of the edge weight (\textit{e.g.}, GAT \cite{velivckovic2018graph}, FAGCN \cite{bo2021beyond} and GBK-GNN \cite{du2022gbk}), we follow the Eq. \eqref{eq:reweight} to combine the LLM-based weight and graph-based weight to perform fine-grained context aggregation of GNN.

\section{Hyper-parameters}
\label{app:hyper}
We run all the experiments on an NVIDIA A800 GPU. For LLM4HeG, the edge weight margin $\alpha=0.3$ and the regularization coefficient $\lambda=0.1$. For the baselines, we use the hyper-parameters as listed in previous literature. For the number of hops in the $i$-hop neighborhood of each node, we use a 1-hop and 2-hop neighborhood for H2GCN \cite{zhu2020beyond} and a 1-hop neighborhood for the other backbones. The hidden unit for GCN and GAT is 16. The number of heads in GAT is 3 for Amazon and 8 for other datasets. For H2GCN, we adopt the H2GCN-1 variant using one embedding round ($K=1$). The parameter setting of FAGCN is: the hidden unit = 32, layers = 2, $\epsilon = 0.4$. For JacobiConv, the parameter setting is $\gamma = 2$ for Polynomial Coefficient Decomposition (PCD), $a=0.5$ and $b=0.25$ for Jacobi Basis. For OGNN, the number of MLP layers is 1 for Cornell, Texas and Wisconsin and 2 for the Actor and Amazon dataset. For SEGSL, the height of the encoding tree $K=2$. The subtree sampling parameter $\theta$ is 2 for the Amazon dataset and 3 for other datasets. The GNN encoder model for the reconstructed graph is GraphSAGE. For DisamGCL, the weight of historical memory $\mu = 0.6$, the controlling variables for the node similarity $\epsilon_1 = 0.74$ and $\epsilon_2$ = 0.4, the threshold of the node similarity $\mathcal{T} = 0.8$, the number of augment instances $K=8$, the weight of contrastive loss $\lambda=1$.

\begin{figure}[tbp] 
\centering
    \centering
    \includegraphics[scale=0.7]{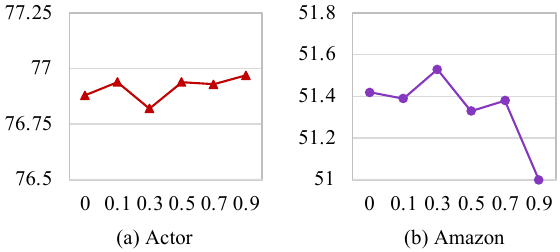}
\caption{Effect of the edge weight margin $\alpha$.}
\label{app: alpha}
\end{figure}

The effect of edge weight margin $\alpha$ of LLM4HeG for other datasets is shown in Fig. \ref{app: alpha}. Generally speaking, the results in Fig. \ref{hyper: alpha} and Fig. \ref{app: alpha} show that $[0.3,0.5]$ appears to be a good range for $\alpha$ on all datasets.

\section{More Details for Baselines}
\label{app:baselines}
\begin{itemize}[leftmargin=*]
\item H2GCN \cite{zhu2020beyond} considers higher-order neighbors, ego-neighbor embedding separation and intermediate layer representations for heterophilic graph.

\item FAGCN \cite{bo2021beyond} employs a self-gating mechanism to adaptively integrate low- and high-frequency signals during message passing.

\item  JacobiConv  \cite{wang2022powerful} deserts nonlinearity and approximates filter functions with Jacobi polynomial bases. 

\item GBK-GNN \cite{du2022gbk} introduces a learnable kernel
selection gate to discriminate node pairs and apply two different kernels for homophily and heterophily node pairs.

\item  OGNN \cite{songordered} introduces an ordered gating mechanism for message passing, effectively handling heterophily and mitigating the over-smoothing problem.

\item  SEGSL \cite{zou2023se} is a graph structure learning framework leveraging structural entropy and the encoding tree to improve both the effectiveness and robustness.

\item  DisamGCL \cite{zhao2024disambiguated} automatically identifies ambiguous nodes and dynamically augments the learning objective through a contrastive learning framework.
\end{itemize}

\balance 
\section{Comparison to Direct LLM Predictions}
\label{app:LLM}
Given the rich textual data associated with the nodes, it is possible to directly feed this information into the LLM for classification, bypassing the use of a GNN. 
We conduct experiments using the original and fine-tuned LLMs.

(1) ``LLM-nofinetune'' employs the original Vicuna 7B model to make the prediction, with the following prompt template for the Cornell, Texas, and Wisconsin datasets. A similar prompt template with different background and task descriptions is used for the Actor and Amazon datasets.\\

\noindent\framebox{\parbox{0.97\linewidth}{\small
\emph{Background:} I have a dataset containing web page information collected from computer science department websites of various universities. These web pages have been manually categorized into five categories, including student, staff, faculty, course, and project.\\

\emph{Task:} I will provide you with the text information of a web page, and I would like you to classify it into one of the following categories: student, staff, course, faculty, or project.\\

The web page content: <text>\\

You may only output the category name, and do not discuss anything else!
}}
\\[2mm]

(2) ``LLM-finetune'' employs the fine-tuned Vicuna 7B model using the edge discriminator in Stage 1 to make the prediction, using the same prompt template as above.

As shown in Table~\ref{appt: LLM}, the performance of the original LLM without fine-tuning performs poorly when directly provided with textual information. This is likely because while LLMs are powerful for natural language processing tasks, they struggle to infer node categories without adaptation to the specific task. After fine-tuning with the LLM-enhanced edge discriminator in Stage 1, the performance improves significantly, but is still worse than LLM4HeG. This highlights the importance of combining node semantic features, structural contexts, and LLM-inferred edge characteristics for effective node classification.
\begin{table}[tbp]
    \centering
    \small
\setlength{\tabcolsep}{2pt}
    \begin{tabular}{lccccc}
        \hline
        & Cornell & Texas & Wisconsin & Actor & Amazon \\ 
        \hline
        LLM-nofinetune & 26.16 & 25.00 & 26.32 & 24.75 & 35.44 \\ 
        LLM-finetune   & 61.90 & 40.91 & 71.93 & 59.96 & 36.52 \\ 
        \textbf{LLM4HeG} & \textbf{75.95} & \textbf{87.50} & \textbf{84.91} & \textbf{76.54} & \textbf{51.53} \\ 
        \hline
    \end{tabular}
    \vspace{-0.2cm}
    \caption{Performance comparison with LLMs.}
    \label{appt: LLM}
\end{table}

\end{document}